\documentclass[runningheads]{llncs}
\usepackage{times}
\usepackage{epsfig}
\usepackage{graphicx}
\usepackage{amsmath}
\usepackage{amssymb}
\usepackage{booktabs}
\usepackage{multirow}
\begin{document}
\title{Multi-scale Neural ODEs for 3D Medical Image Registration}
\titlerunning{Neural ODEs for Image Registration}
%
\author{Junshen Xu\thanks{This work was carried out during the internship of the author at United Imaging Intelligence, Cambridge, MA 02140.}\inst{1} \and
Eric Z. Chen\inst{2} \and
Xiao Chen \inst{2} \and
Terrence Chen\inst{2} \and
Shanhui Sun\inst{2}
}

\authorrunning{J. Xu et al.}

\institute{Department of Electrical Engineering and Computer Science, MIT, \\ Cambridge, MA, USA \\\email{junshen@mit.edu} \and
United Imaging Intelligence, Cambridge, MA, USA
\\\email{\{zhang.chen, xiao.chen01, terrence.chen, shanhui.sun\}@united-imaging.com}
}

\maketitle              
\setcounter{footnote}{0}

\begin{abstract}
Image registration plays an important role in medical image analysis. Conventional optimization based methods provide an accurate estimation due to the iterative process at the cost of expensive computation. Deep learning methods such as learn-to-map are much faster but either iterative or coarse-to-fine approach is required to improve accuracy for handling large motions. In this work, we proposed to learn a registration optimizer via a multi-scale neural ODE model. The inference consists of iterative gradient updates similar to a conventional gradient descent optimizer but in a much faster way, because the neural ODE learns from the training data to adapt the gradient efficiently at each iteration. Furthermore, we proposed to learn a modal-independent similarity metric to address image appearance variations across different image contrasts. We performed evaluations through extensive experiments in the context of multi-contrast 3D MR images from both public and private data sources and demonstrate the superior performance of our proposed methods.
\keywords{Multi-modal image registration \and Neural ordinary differential equations \and Disentangled representation \and Self-supervised learning.}
\end{abstract}

\section{Introduction}
Image registration is an essential step in many tasks such as motion correction and atlas-based image segmentation. 
It is indispensable in many clinical applications such as surgical planning \cite{risholm2011multimodal} and radiogenomics analysis~\cite{incoronato2017radiogenomic}, where 3D images are commonly used. 
Conventional image registration methods solve an optimization problem by minimizing the dissimilarity between the transformed image and the target image. 
Although the conventional methods often achieve high accuracy, the slow process and expensive computation due to the use of iterative non-linear optimization algorithms hinder their clinical translation. 

Recently deep learning based approaches have been proposed for image registration \cite{balakrishnan2019voxelmorph,de2017end,rohe2017svf,cao2017deformable}, which learn to map from image or feature space to a spatial transformation space using neural networks trained on large datasets. 
Since the registration during the inference is just one forward pass of the network, the deep learning methods are intrinsically faster than the conventional methods. Supervised methods ~\cite{cao2017deformable,rohe2017svf,yang2017quicksilver} require ground truth transformations for training, which are typically difficult to obtain in clinical practice especially for deformable registration.  Unsupervised methods, such as DIRNet~\cite{de2017end} and VoxelMorph~\cite{balakrishnan2019voxelmorph}, directly regress deformation fields by minimizing dissimilarity between input and target images. 
However, for large motion, the learn-to-map based methods often do not perform well~\cite{shen2019networks}. 
Multi-stage methods are proposed~\cite{de2019deep,zhao2019recursive}, where several networks are cascaded to gradually refine the estimated transformation. Deep reinforcement learning is also applied to image registration, especially for rigid registration~\cite{ma2017multimodal,hu2020end,sun2018robust}. In terms of non-rigid registration, Krebs \textit{et al.}~\cite{krebs2017robust} proposed an agent-based method for prostate MR registration, which is limited to low dimensional B-spline.

Multi-modal images are frequently used in clinic. The main challenge of multi-modal image registration is to find a proper dissimilarity cost function to distinguish motions from contrast changes. 
Some classical methods use mutual information (MI)~\cite{maes1997multimodality,de2010hierarchical} and modality-independent neighborhood descriptor (MIND)~\cite{heinrich2012mind}.  
Some learning-based methods convert the problem to mono-modal registration by image-to-image translation~\cite{cao2017dual,arar2020unsupervised}, which are prone to synthetic artifacts.
UMDIR~\cite{qin2019unsupervised} measured the difference between multi-modal images in a feature space; however, only 2D deformable image registration between two modalities was addressed.

Recently, neural ordinary differential equations (ODEs) are proposed to represent more complex dynamics over classical ODEs. Compared to the common deep learning models such as ResNet and UNet, neural ODE models are more memory and parameter efficient and provide the benefits of adaptive computation~\cite{chen2018neural}, which are potentially suitable for medical applications~\cite{chen2020mri}. 
The optimization dynamics are also inherently continuous. These merits motivate us to learn the optimization in medical image registration using neural ODEs.
To the best of our knowledge, our study is the first work to apply neural ODEs to image registrations.

The contributions of our proposed methods are summarized as follows: 
1) We proposed a new direction of modeling image registration optimizer as a continuous optimization dynamics via neural ODEs. 
2) We introduced multi-scale architecture to neural ODEs to reduce searching space by performing registration iteratively on different scales.
3) Our proposed method is a general learn-to-learn image registration framework and is not limited to specific transformations. 
4) Our framework can handle multiple contrasts with a single trained network attribute to proposed contrast-independent similarity metric for $n (\ge 2)$ modalities. 

\section{Method}
Let $x_\text{mov}$ and $x_\text{fix}$ denote the moving and fixed images, respectively, and let $\phi_\theta$ be the transformation field between two images parameterized by $\theta$. The image registration optimization problem can be written as:
\begin{equation}
    \hat{\theta} = \arg\min_\theta \mathcal{L}(x_\text{mov}\circ\phi_\theta, x_\text{fix}) + \mathcal{R}(\theta),
    \label{eqn:opt}
\end{equation}
where $x_\text{mov}\circ\phi_\theta$ is the transformed image, $\mathcal{L}$ is the dissimilarity cost function, and $\mathcal{R}$ is the regularization term. The form of $\theta$ is determined by the types of transformation. 

\begin{figure*}[t]
\begin{center}
  \includegraphics[width=0.8\linewidth]{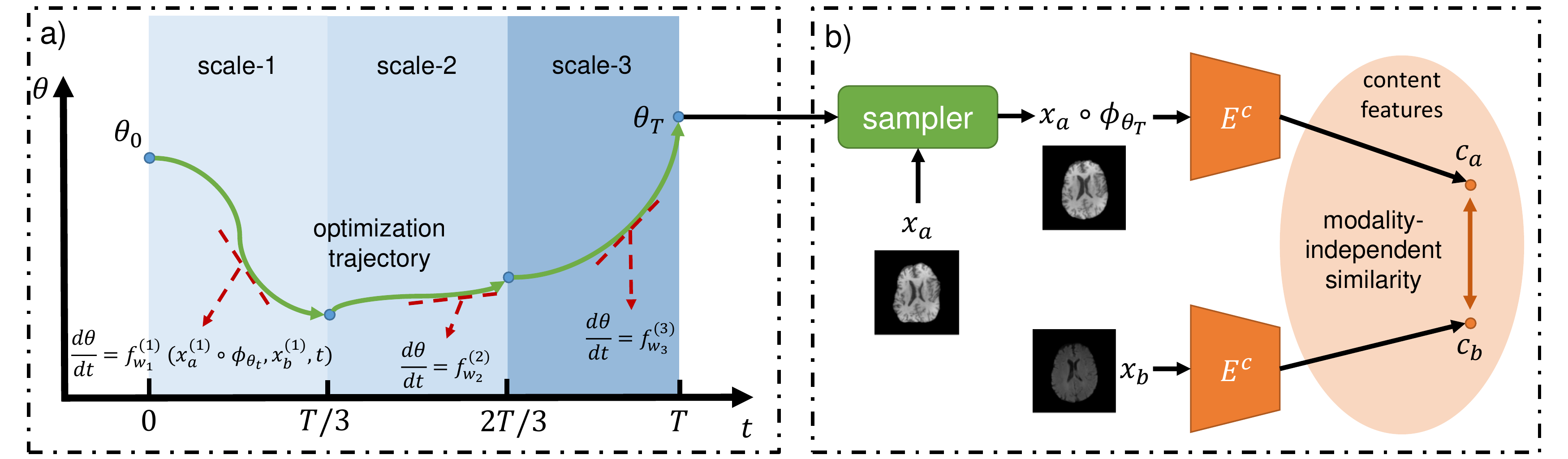}
\end{center}
  \caption{Overview of our proposed method: a) Image registration optimization is modeled as a continuous optimization dynamics via neural ODEs, b) A modality-independent similarity metric is realized via a pretrained content encoder.}
\label{fig:overview}
\end{figure*}

We propose a novel multi-modal image registration method by learning an image registration optimizer via neural ODEs, and deriving loss function from a pretrained image content encoder. Fig. \ref{fig:overview} illustrates an overview of our method. 

\subsection{Learn registration optimizer via neural ODEs}

The optimization problem for image registration in Eq. \ref{eqn:opt} can be solved via gradient descent optimizer. Thus, the update of $\theta$ at each step is $\theta_{t+1} = \theta_{t} - \eta_t {\partial \mathcal{(L+R)}}/{\partial\theta_t}\triangleq\theta_t + f(\theta_t, t)$, 
where $t$ represents $t^{th}$ step and $\eta_t$ is the step size. If the time difference is small enough, the difference equation can be re-written as an ODE function, $\frac{d\theta_t}{dt} = f(\theta_t, t), t\in [0, T]$. 
Given the initial parameter $\theta_0$, the final parameter $\theta_T$ is the solution to this ODE initial value problem. 
Some conventional methods such as LDDMM~\cite{beg2005computing} and SyN~\cite{avants2008symmetric} describe the evolution of transformation as a differential equation. These methods solve differential equations where system dynamics are described by predefined functions, which is less flexible. 
Neural ODEs use trainable network to replace the predefined function, which can be considered as learning an optimizer to compute the gradient update. $\theta_T$ is the inference of the neural ODE model. 
We further assume that for two 3D images $x_a$ and $x_b$, the evolution of $\theta_t$ follows a neural ODE:
\begin{equation}
    \frac{d\theta_t}{dt}=f_w(x_{a}\circ\phi_{\theta_t}, x_{b}, t), t\in[0,T],
\end{equation}
where $f_w$ is a neural network with parameter $w$ which takes the current warped image $x_{a}\circ\phi_{\theta_t}$, the target image $x_b$ and the time variable $t$ as inputs. Therefore, the final output at time $T$ can be computed by integrating function $f$ over time interval $[0, T]$. 
In practice, the integral is evaluated by an ODE solver, such as Runge–Kutta methods and adaptive step size solver. To train the neural ODE model, we adopt the adaptive checkpoint adjoint method~\cite{zhuang2020adaptive}.

\begin{figure*}[t]
\begin{center}
\includegraphics[width=0.9\linewidth]{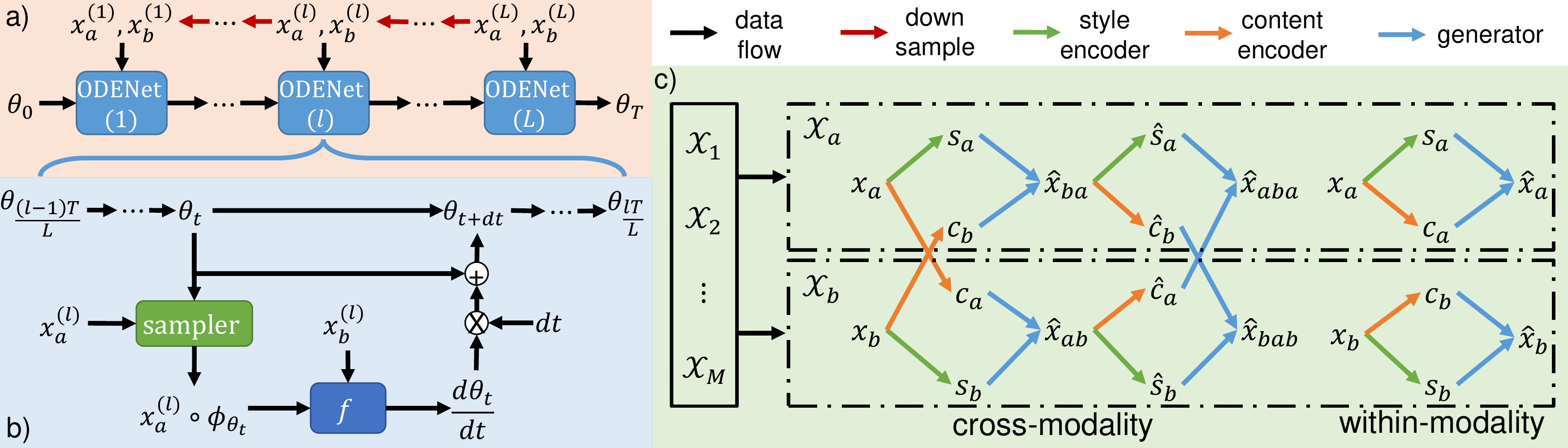}
\end{center}
  \caption{Implementation of our proposed framework: 
  a) Overview of multi-scale ODENet: several cascaded ODENets in different resolutions,
  b) a computational graph of a single scale ODENet with Euler's solver, 
  c) overview of the utilized image translation towards metric learning.}
\label{fig:network}
\end{figure*}

{\bf Multi-scale ODE network: }Empirically, we found that solving neural ODE for image registration problem requires an extensive number of function evaluations (NFE) by ODE solver, which leads to prolonged training and inference time. To address this problem, we propose a multi-scale ODE network (MS-ODENet) which performs registration by solving ODEs at different resolutions. Specifically, let $\{x_i^{l}\}_{l=1}^L (i=a, b)$ be an image pyramid with $L$ different resolutions, where $x^{(L)}_i=x_i$, and $x^{(l-1)}_i$ is generated by down sampling $x^{(l)}_i$ with a factor of 2 on all 3 axes. We divide the whole time interval $[0, T]$ into $L$ congruent segments. In each segment, we solve the ODE in Eq.~\ref{eqn:msode} at the corresponding resolution as shown in Fig. \ref{fig:network} a and b.
\begin{equation}
\label{eqn:msode}
    \frac{d\theta_t}{dt}=f^{(l)}_{w_l}(x_{a}^{(l)}\circ\phi_{\theta_t}, x_{b}^{(l)}, t), t\in[\frac{l-1}{L}T,\frac{l}{L}T],
\end{equation}
where $f^{(l)}_{w_l}$ is the network at the $l$-th scale. The output parameter $\theta_T$ is the integral over all scales. 
The benefits of this design are two-fold: 1) The time cost for function evaluations is much smaller at low resolutions, allowing a larger number of steps to reach the desired accuracy. 2) The searching space for image registration is largely reduced and therefore the convergence of the ODE network is faster, which also makes the model less sensitive to local optimal. 

{\bf Loss functions: }The proposed registration network is trained in an unsupervised manner by minimizing a loss function similar to Eq. \ref{eqn:opt}. The utilized similarity metric $\mathcal{L}_\text{sim}$ (Fig.~\ref{fig:overview} b) between two images ($x_{a}$ and $x_{b}$) is defined in Eq.~\ref{eqn:similarity}:
\begin{equation}
    \mathcal{L}_{\text{sim}} =\mathbb{E}_{x_a,x_b,a,b} ||E^c_{3D}(x_{a}\circ\phi_{\theta_T}) - E^c_{_{3D}}(x_{b}) ||_2^2,
    \label{eqn:similarity}
\end{equation}
where $E^c_{_{3D}}(\cdot)$ is a 3D content feature extractor.  The 3D content feature is composed of 2D content features generated from $N$ randomly selected 2D images of different axes from the image utilizing a 2D feature extractor $E^c$ (Section~\ref{sec:metrics}).
In addition, we perform a random perturbation to a given image $x_a$ such that $\Tilde{x}_a=x_a\circ\phi_{\Tilde{\theta}}$, where $\Tilde{\theta}$ is randomly sampled from a distribution of parameter $\mathbb{P}_\theta$. The pair $x_a$ and $\Tilde{x}_a$ are fed to the registration network resulting transformation parameters $\Tilde{\theta}_T = \text{MS-ODENet}(\theta_0, x_a, \Tilde{x}_a, L, T)$. 
Our network can align two images from the same modality so that we expect $\Tilde{\theta}_T$ approximating to the purturbation $\Tilde{\theta}$ by minimizing L2 loss: $\mathcal{L}_{\text{self}}=\mathbb{E}_{x_a,a,\Tilde{\theta}}||\Tilde{\theta}-\Tilde{\theta}_T||_2^2$. The total loss function is summarized as $\mathcal{L} = \mathcal{L}_\text{sim} + \lambda_\text{self} \mathcal{L}_\text{self} + \lambda_\text{reg} \mathcal{L}_{\text{reg}}$, 
where $\mathcal{L}_{\text{reg}}= \mathbb{E}_{x_a,x_b,a,b}||\nabla\phi_{\theta_T}||_2^2$ is the regularization term enforcing a smooth motion field in deformable registration. $\lambda_\text{self}$ and $\lambda_\text{reg}$ are weighting coefficients.

\subsection{Pretrained Feature Extraction network}
\label{sec:metrics}
Different contrast images from the same patient share the same anatomical structures (content features) but have different style features. This inspires us decomposing the image into content features and style features in the latent space. The realization of this feature domain disentanglement is extended from the diverse image-to-image translation framework~\cite{lee2020drit++}. We trained content encoder $E^c$, style encoder $E^s$ and generator $G$ to perform image translation from one contrast to another. Note that only $E^c$ is used in our registration task. Fig.~\ref{fig:network}c illustrates an overview of our feature extraction framework utilizing image-to-image translation. 
 Suppose two different groups $\mathcal{X}_a$ and $\mathcal{X}_b$ sampled from $M$ modality groups ($\{\mathcal{X}_i\}_{i=1}^M$). Given $x_a$ and $x_b$ two image samples in these two groups, we perform cross-modality translation as follows: (a) Extract the content and style features, i.e., $s_a= E^s(x_a, a)$, $c_a=E^c(x_a)$, $s_b= E^s(x_b, b)$, $c_b=E^c(x_b)$. (b) Swap style features and generate translated images, i.e., $\hat{x}_{ab} = G(c_a, s_b, b)$, $\hat{x}_{ba}=G(c_b, s_a, a)$. (c) Encode the translated images to reconstruct content and style features of the original images, $\hat{s}_a = E^s(\hat{x}_{ba}, a)$, $\hat{c}_a = E^c(\hat{x}_{ab})$, $\hat{s}_b = E^s(\hat{x}_{ab}, b)$, $\hat{c}_b = E^c(\hat{x}_{ba})$. (d) Swap the style features again to generate the original images, $\hat{x}_{aba} = G(\hat{c}_a, \hat{s}_a, a)$, and $\hat{x}_{bab} = G(\hat{c}_b, \hat{s}_b, b)$. We trained the network using adversial loss in ~\cite{lee2020drit++}. Moreover, we expect that $\hat{x}_{aba}$ and $\hat{x}_{bab}$ are consistent to the original images $x_a$ and $x_b$ respectively. We use L1 loss $\mathcal{L}_{\text{cyc}}$ to enforce this cycle consistency. We also introduce the feature reconstruction loss for content/style features, $\mathcal{L}_{\text{rec}}^{c}$/$\mathcal{L}_{\text{rec}}^{s}$, which are the L1 loss between $c_a, c_b$/$s_a, s_b$ and $\hat{c}_a, \hat{c}_b$/$\hat{s}_a, \hat{s}_b$. Similarly, we define mono-modal reconstruction loss $\mathcal{L}_{\text{rec}}^{x}$ as the L1 loss between the reconstructed images $\hat{x}_a = G(c_a, s_a, a)$, $\hat{x}_b=G(c_b, s_b, b)$ and the original images $x_a, x_b$. Since training is not trivial in 3D due to high dimensionality, we use a 2D multi-channel (adjacent slices in 3D volume) network.

\section{Experiments and Results}

\subsection{Experiment setup}
{\bf Dataset: } Currently, there is a lack of large scale medical image dataset dedicated to multi-modal image registration. We use a public dataset and a separate acquired volunteer dataset. The public dataset is the brain tumor segmentation (BraTS) 2020 dataset~\cite{menze2014multimodal} which consists of multi-modal 3D brain MRI  with four distinctive contrasts (T1, T2, T2-FLAIR, and T1Gd) from 494 subjects with glioblastomas, resulting in various multi-modal image registration tasks (12 pairs per subject). Besides, tumor masks provide a clinically meaningful evaluation metric for registration. For each subject, all modalities were normalized into 3D volumes with a size of $240\times240\times160$ and a resolution of $1\times1\times1\text{mm}^3$. The dataset is split into 444 subjects (5,328 pairs) for training and validation and 50 subjects (600 pairs) for testing. Since different modalities have been registered, we simulate motions by applying random rigid transformation, random control point deformation, or both. 
Note that the simulated transformation fields are only used for evaluation, not for training. We also acquired additional multi-modal 3D MR brain data on 25 volunteers using a special MR technique that can acquire multiple contrasts simultaneously. 
This private dataset is used only for testing the generalizability of the proposed methods. The motion is simulated as described above. Images were preprocessed to remove skull using DeepBrain\footnote{\url{https://github.com/iitzco/deepbrain}} and resized as the public data.  

\begin{table}[t]
\scriptsize
\caption{Quantitative results for image registration on BraTS, where R, D, and B represent rigid, dense and B-spline parameterization in MS-ODENet. The mean (standard deviation) are reported.}
\begin{center}
\begin{tabular}{c|c|cccc}
\toprule
Transformations & Methods & Dice / \% & RMSE($x$)$\times10^{-2}$ & RMSE($\phi$) / mm & Time / s \\
\midrule
\multirow{3}{*}{rigid} &ANTs &  63.0 (40.5) & 8.34 (7.64) & 7.28 (6.66) & 17.17 (4.05) \\
 &ANTs+I2I &  60.6 (40.8) &  8.83 (7.81) & 7.54 (6.78) & 30.69 (4.70) \\
 &MS-ODENet(R) & \bf{90.6 (17.2)} & \bf{3.89 (2.95)} & \bf{3.57 (5.18)} & \bf{0.55 (0.08)} \\
\midrule
\multirow{6}{*}{deformable} &ANTs &  81.9 (8.8) & 6.31 (1.98) & 1.21 (0.37) & 55.35 (7.07) \\
&ANTs+I2I &  81.1 (8.3) &  6.34 (1.82) & 1.06 (2.35) & 68.47 (6.18) \\
&VM  &  79.4 (8.7) &  8.81 (3.03) & 1.61 (0.69) & \bf{0.24 (0.05)} \\
&VM+I2I &  80.1 (7.8) &  8.52 (2.30) & 1.26 (0.31) & 0.34 (0.06) \\
&MS-ODENet(D) & 81.6 (8.1) & 6.63 (2.22) & 1.11 (0.18) & 1.13 (0.10) \\
&MS-ODENet(B) & \bf{83.0 (8.7)} & \bf{6.17 (2.43)} & \bf{0.99 (0.32)} & 0.31 (0.06) \\
\midrule
\multirow{7}{*}{rigid+deformable} &ANTs &  73.6 (22.2) & 6.79 (2.51) & 2.99 (2.95) & 70.87 (7.97) \\
&ANTs+I2I &  71.1 (22.1) &  6.97 (2.26) & 2.83 (2.80) & 87.25 (12.2) \\
&RCN & 64.9 (10.1) & 8.74 (3.21) & 5.48 (1.97) & 2.49 (0.31) \\
&VM$^*$  &  78.1 (9.8) &  8.62 (3.02) & 2.92 (2.31) & 0.60 (0.23) \\
&VM$^*$+I2I &  78.4 (9.2) &  7.07 (1.96) & 2.04 (1.38) & 1.01 (0.38) \\
&MS-ODENet(R+D) & 79.6 (9.1) & 6.70 (2.28) & 1.82 (1.22) & 1.39 (0.12) \\
&MS-ODENet(R+B) & \bf{81.1 (9.9)} & \bf{6.28 (2.38)} & \bf{1.52 (1.02)} & \bf{0.56 (0.09)} \\
\bottomrule
\end{tabular}
\end{center}
\label{tab:quantatitive_all}
\end{table}

{\bf Registration networks:} We evaluated our methods on rigid, deformable and a hybrid of rigid and deformable motions. For 3D rigid motion, $f$ includes convolution layers followed by fully connected layers. The output size of $f$ is the same as the degree of freedom in the transformation. For deformable motion, fully convolutional network is used. One variation of the deformable registration is kernel based method (B-spline). Parameter $\theta$ is the grid of control points which can also be regarded as an image. For dense motion, the UNet as in ~\cite{ronneberger2015u} is used for voxel-wise estimation. For hybrid motion, we cascade the rigid and deformable MS-ODENets sequentially.

{\bf Feature extraction network:} 
The backbone network is similar to \cite{huang2018multimodal}. The content encoder is a fully convolutional network while the style encoder has convolution layers followed by a global average pooling and a fully connected layer, which forces the network to extracts global style features. The generator is a CNN with deconvolution layers to generate images of the original size. Besides, to condition on modality $\mathcal{X}_i$ for style encoder and image generator, the modality code $i$ is converted into a one-hot vector concatenated with the input tensor along the channel dimension.

{\bf Implementation details: }We follow the protocol in~\cite{arjovsky2017wasserstein} for training GAN with number of slices set to 5. For registration, we set $\lambda_\text{self}=1$, and let $\lambda_\text{reg}$ be 10 and 2 for dense and B-spline respectively. We use $L=T=3$ for MS-ODENet. Let $F(dt)$ be Euler's method with fixed step size $dt$ and $A(\epsilon)$ be adaptive Heun's method with tolerance of error $\epsilon$. 
We use adaptive solvers when the search space is small (Rigid: $A(10^{-3})$-$A(10^{-3})$-$F(0.1)$) to avoid extensive NFE and use fixed step size solver for large search space (B-spline: $F(0.2)$-$F(0.2)$-$F(0.25)$, Dense: $F(0.2)$-$F(0.2)$-$F(0.5)$).
All networks are trained using Adam~\cite{kingma2014adam} optimizer with a learning rate of $10^{-4}$ on an NVIDIA Tesla V100 GPU.

{\bf Evaluation metrics: }For evaluation, Dice scores \cite{dice1945measures} are computed over tumor masks. With available synthetic transformation fields and ground truth images, root mean square errors are calculated and denoted as RMSE($\phi$) and RMSE($x$), respectively. The metrics are averaged over all pairs of test data.

\subsection{Results}
\label{sec:transformation_variation}
\begin{figure*}[t]
\begin{center}
   \includegraphics[width=0.9\linewidth]{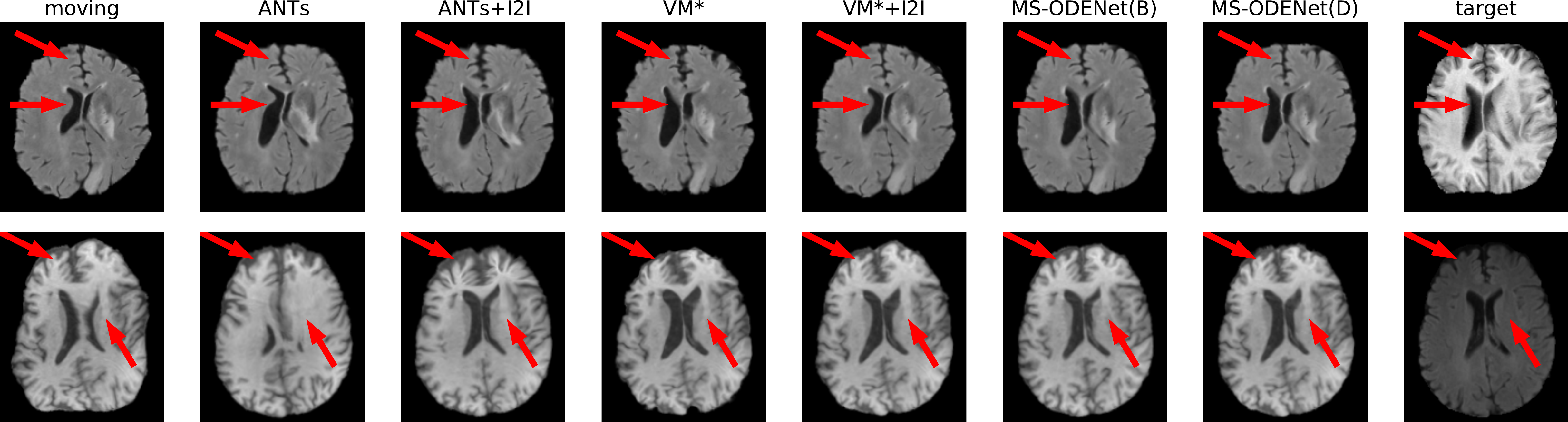}
\end{center}
   \caption{Examples of compared methods on deformable registration. Top row: T2-FLAIR to T1 registration. Bottom row: T1 to T2-FLAIR. Red-arrows highlight example different areas.}
\label{fig:deform}
\end{figure*}

Table~\ref{tab:quantatitive_all} shows the quantitative results for rigid, deformable and hybrid registration (rigid+deformable). 
{\bf Rigid:}~Random rotation and translation were synthesized along all three axes and were sampled from $U(-75^{\circ}, 75^{\circ})$ and $U(-20,20)\mbox{mm}$, respectively. We used the rigid registration method with MI metric in Advanced Normalization Tools (ANTs) \cite{avants2009advanced} as the baseline. We also used the trained GAN to perform image translation followed by mono-modal image registration with ANTs (ANTs+I2I). 
{\bf Deformable:} 
We made synthetic image pairs through elastic transformations by perturbing B-spline control points with noise from $\mathcal{N}(0,(8\text{mm})^2)$ on three axes. For comparison, we use SyN~\cite{avants2008symmetric} in ANTs, VoxelMorph (VM)~\cite{balakrishnan2019voxelmorph} with MI metric, and also their variants with image translation (ANTs+I2I and VM+I2I). For MS-ODENet, we use two different parameterizations, namely dense deformation (D) and B-spline (B). Fig.~\ref{fig:deform} shows example results. 
{\bf Rigid + deformable:} 
Random rotations, translations and control point noise are from $U(-40^\circ,40^\circ)$, $U(-10,10)\mbox{mm}$, and $\mathcal{N}(0,(8\text{mm})^2)$, respectively. We compared our method to SyN, RCN~\cite{zhao2019recursive} and VM. RCN is an iterative deep learning method consisting of an affine network and $n$ deforamble networks. We use MI metric for training RCN and set $n=2$ due to memory limit. Since VM was proposed to solve deformable registration, we performed a rigid registration using our rigid MS-ODENet prior to applying VM. This approach is denoted as VM$^*$.

\begin{figure}[t]
\begin{center}
   \includegraphics[width=1.0\linewidth]{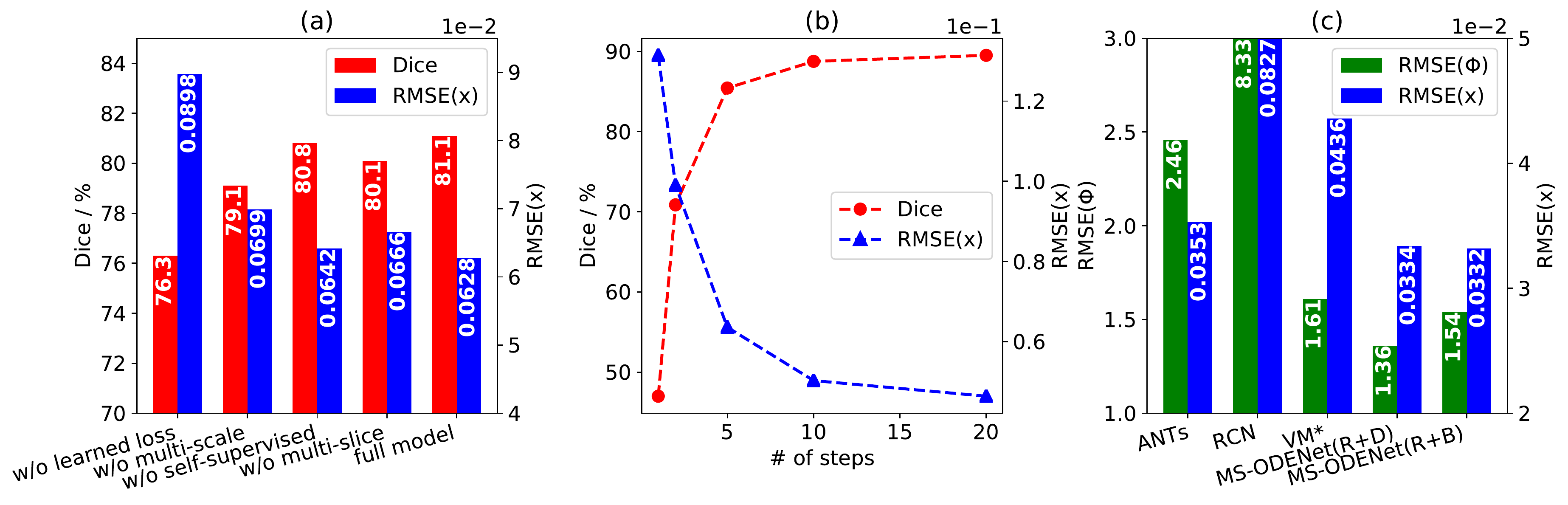}
\end{center}
   \caption{a) Quantitative results for the ablation study. b) Performance on large motion for neural ODE models with different number of solver steps. c) Results for the generalizability study.}
\label{fig:new_ablate}
\end{figure}

{\bf Ablation:}~Fig.~\ref{fig:new_ablate}(a) summarizes the results that evaluate models with no learned content loss (replaced by MI), no multi-scale ODE, no self-supervision or no multi-slice GAN respectively and compare them with the full model on rigid + B-spline deformable registration. 
To investigate the necessity of iterative registration for large motions, we conducted a rigid registration experiment with $L=T=1$ and various numbers of steps in ODE solver. Fig. \ref{fig:new_ablate}(b) shows the corresponding results.

{\bf Generalization:} We performed the rigid+deformable test on our private dataset. 
Fig.~\ref{fig:new_ablate}(c) shows bar chart among compared methods. 

\section{Discussions and Conclusions}
Table~\ref{tab:quantatitive_all} shows that our proposed MS-ODENet outperforms classical methods under various transformations. In all experiments, MS-ODENet is much faster than ANTs due to the fast inference of neural networks and the smaller number of iterations needed in neural ODEs. In rigid registration (Fig.~\ref{tab:quantatitive_all}), MS-ODENet greatly outperforms ANTs with or without the domain translation. Classical methods only consider local gradient information and tend to get stuck at local minima. The MS-ODENet learns the optimization process via neural ODEs, which utilizes not only the current local information but also the experiences learned from the dataset, and is thus more likely to reach the global minimum. In deformable registration, our proposed methods achieve similar or better accuracy compared with classical methods. For rigid+deformable registration, ANTs suffers greatly from the additional rigid transformation, indicating that traditional optimization-based methods rely heavily on the initialization of parameters. 

Our methods also outperform other deep learning methods (Table~\ref{tab:quantatitive_all}) on deformable registration. The iterative updates in MS-ODENet improve registration accuracy progressively (Fig.~\ref{fig:new_ablate}(c)). When the step number is one, the MS-ODENet is equivalent to a conventional deep learning model. Note that all the ablated methods (Fig.~\ref{fig:new_ablate}(a)) outperform the other deep learning methods. Furthermore, with the adjoint method, the training of our neural ODE model does not backpropagate through the operations of the solver~\cite{chen2018neural,zhuang2020adaptive} and thus is more memory efficient than traditional deep learning models. Unlike other coarse-to-fine methods that need to be trained in separate stages~\cite{de2019deep}, our multi-scale neural ODE model can be trained end-to-end. In the generalization study, the proposed MS-ODENets show consistent improvement over the other methods. The RCN does not employ iterative networks for affine transformation and therefore has poor generalizability for large transformation.

In this work, we present a new framework for 3D multi-modal image registration. We formulate the optimization in conventional registration methods as a continuous process and learn the optimizer via a neural ODE. Furthermore, for efficient learning and inference, we propose a multi-scale architecture to narrow the searching space from coarse to fine image resolutions. In addition, we employ an image-to-image translation GAN to learn a modality-independent metric between images from different modalities. Experiment results show that our proposed framework is superior to other compared methods. For future work, we will extend our framework to other types of medical registration such as 3D-2D image registration.

%
%
%
\bibliographystyle{splncs04}
\bibliography{egbib.bib}

\end{document}


\title{Supplementary Material}
\titlerunning{Supplementary Material}

\author{Junshen Xu\inst{1} \and
Eric Z. Chen\inst{2} \and
Xiao Chen \inst{2} \and
Terrence Chen\inst{2} \and
Shanhui Sun\inst{2}
}

\authorrunning{J. Xu et al.}

\institute{Department of Electrical Engineering and Computer Science, MIT, \\ Cambridge, MA, USA \\\email{junshen@mit.edu} \and
United Imaging Intelligence, Cambridge, MA, USA
\\\email{\{zhang.chen, xiao.chen01, terrence.chen, shanhui.sun\}@united-imaging.com}
}

\maketitle              
\setcounter{footnote}{0}

\section{Details of Network Implementation}
\subsection{Registration networks}

{\bf Rigid registration:} The architectures of network used for rigid registration are shown in Table \ref{tab:rigid_network}. We use leaky ReLU activation~\cite{maas2013rectifier} with $\alpha=0.2$ for 3D convolutions and ReLU activation for fully connected layers, except the output layer.

\begin{table}[h]
\caption{Network architectures for rigid registration at different scales. c: number of output channels, k: kernel size, s: stride, p: padding, and the numbers after FC denote numbers of output in fully connected layers.}
\begin{center}
\begin{tabular}{c|ccc}
\hline
 & full-scale & 2x downsampled & 4x downsampled \\
\hline
conv3d-1 & c16k4s2p1 & c16k4s2p1 & c16k4s2p1\\
conv3d-2 & c32k4s2p1 & c32k4s2p1 & c32k4s2p1\\
conv3d-3 & c64k4s2p1 & c64k4s2p1 & c64k4s2p1\\
conv3d-4 & c128k4s2p1 & c128k4s2p1 & c128k4s2p1\\
conv3d-5 & c128k4s2p1 & c128k4s2p1 & \\
conv3d-6 & c128k4s2p1 & & \\
\hline
\multicolumn{4}{c}{flatten}\\
\hline
fc-1 & c128 & c128 & c128\\
fc-2 & c6 &  c6 & c6\\
\hline
\end{tabular}
\end{center}
\label{tab:rigid_network}
\end{table}

{\bf B-spline registration:} We set the interval of control points to 15 pixels for all different resolutions. Table \ref{tab:bspline_network} shows the architecture of B-spline registration networks. The networks consist of four 3D convolution layers followed by two 3D resblock~\cite{he2016deep} and a $1\times1\times1$ convolution as output layer.

\begin{table}[h]
\caption{Network architectures for B-spline registration at different scales. c: number of output channels, k: kernel size, s: stride, p: padding.}
\begin{center}
\begin{tabular}{c|ccc}
\hline
 & full-scale & 2x downsampled & 4x downsampled \\
\hline
conv3d-1 & c32k4s2p1 & c32k4s2p1 & c32k4s2p1\\
conv3d-2 & c64k4s2p1 & c64k4s2p1 & c64k4s2p1\\
conv3d-3 & c128k4s2p1 & c128k4s2p1 & c128k4s2p1\\
conv3d-4 & c128k4s2p2 & c128k4s2p2 & c128k4s2p2\\
\hline
resblock-1 & c128k3s1p1 & c128k3s1p1 & c128k3s1p1\\
resblock-2 & c128k3s1p1 & c128k3s1p1 & c128k3s1p1\\
\hline
conv3d-5 & c3k1s1p0 &  c3k1s1p0 & c3k1s1p0\\
\hline
\end{tabular}
\end{center}
\label{tab:bspline_network}
\end{table}

{\bf Dense registration:} The 3D UNet-based~\cite{cciccek20163d} architecture is adopted for dense registration, where the details are shown in Table~\ref{tab:dense_network}.

\begin{table}[h]
\caption{Network architectures for dense registration at different scales. c: number of output channels, k: kernel size, s: stride, p: padding.}
\begin{center}
\begin{tabular}{c|ccc}
\hline
 & full-scale & 2x downsampled & 4x downsampled \\
\hline
conv3d-enc1 & c8k5s1p2 & c8k5s1p2 & c8k5s1p2\\
conv3d-enc2 & c16k4s2p1 & c16k4s2p1 & c16k4s2p1\\
conv3d-enc3 & c32k4s2p1 & c32k4s2p1 & c32k4s2p1\\
conv3d-enc4 & c32k4s2p1 & c32k4s2p1 & \\
conv3d-enc5 & c32k4s2p1 &  & \\
\hline
conv3d-dec5 & c32k3s1p1 & & \\
upsample  & 2x &  & \\
concat  & conv3d-enc4  &  & \\
\hline
conv3d-dec4 & c32k3s1p1 & c32k3s1p1 & \\
upsample  & 2x & 2x & \\
concat & conv3d-enc3 & conv3d-enc3 & \\
\hline
conv3d-dec3 & c32k3s1p1 & c32k3s1p1 & c32k3s1p1\\
upsample  & 2x & 2x & 2x \\
concat  &  conv3d-enc2 &  conv3d-enc2 &   conv3d-enc2\\
\hline
conv3d-dec2 & c16k3s1p1 & c16k3s1p1 & c16k3s1p1\\
upsample  & 2x & 2x & 2x\\
concat  & conv3d-enc1 & conv3d-enc1 & conv3d-enc1\\
\hline
conv3d-dec1 & c8k3s1p1 & c8k3s1p1 & c8k3s1p1\\
\hline
conv3d-out & c3k3s1p1 & c3k3s1p1 & c3k3s1p1\\
\hline
\end{tabular}
\end{center}
\label{tab:dense_network}
\end{table}

\subsection{GAN}

{\bf Encoders:} In the proposed multi-slice GAN, the encoders take adjacent slices as input (5 slices in our experiment). Besides, to condition on the modality for style encoder, the modality code $i$ is converted into a one-hot vector concatenated with the input tensor along the channel dimension. Table~\ref{tab:content_network} and~\ref{tab:style_network} show the architectures of content encoder and style encoder respectively.

\begin{table}[h]
\caption{Architectures of content encoder. c: number of output channels, k: kernel size, s: stride, p: padding.}
\begin{center}
\begin{tabular}{c|c}
\hline
layers & hyper-parameters\\
\hline
conv2d-1 & c64k7s1p3\\
conv2d-2 & c128k4s2p1\\
conv2d-3 & c256k4s2p1\\
\hline
resblock-1 & c256k3s1p1\\
resblock-2 & c256k3s1p1\\
resblock-3 & c256k3s1p1\\
resblock-4 & c256k3s1p1\\
\hline
\end{tabular}
\end{center}
\label{tab:content_network}
\end{table}

The content encoder is a fully convolutional network that extract local content features from images while the style encoder consists of 5 convolution layers followed by global average pooling and a fully connected layers to extract global style features which is embedded to an 8 dimensional space. 

\begin{table}[h]
\caption{Architectures of style encoder. c: number of output channels, k: kernel size, s: stride, p: padding.}
\begin{center}
\begin{tabular}{c|c}
\hline
layers & hyper-parameters\\
\hline
conv2d-1 & c64k7s1p3\\
conv2d-2 & c128k4s2p1\\
conv2d-3 & c256k4s2p1\\
conv2d-4 & c256k4s2p1\\
conv2d-5 & c256k4s2p1\\
\hline
\multicolumn{2}{c}{global average pooling}\\
\hline
fc-1 & c8\\
\hline
\end{tabular}
\end{center}
\label{tab:style_network}
\end{table}

{\bf Generator:} The architecture of generator is shown in Table~\ref{tab:generator_network}. The input of generator is the content features concatenated with the one-hot modality code. To further utilize style information, Adaptive Instance Normalization (AdaIN)~\cite{huang2018multimodal} is used in the resblocks.
\begin{equation}
    \text{AdaIN}(x,\gamma,\beta)=\gamma\frac{x-\mu}{\sigma} +\beta
\end{equation}
where $x$ is the activation of previous convolution layer, $\mu$ and $\sigma$ are channel-wise mean and standard deviation, $\gamma$ and $\beta$ are affine parameters. $[\gamma,\beta]=\text{MLP}(s)$, where $s$ is the style content and MLP is a 3-layer perceptron.

\begin{table}[h]
\caption{Architectures of generator. c: number of output channels, k: kernel size, s: stride, p: padding.}
\begin{center}
\begin{tabular}{c|c}
\hline
layers & hyper-parameters\\
\hline
resblock-1 & c260k3s1p1\\
resblock-2 & c260k3s1p1\\
resblock-3 & c260k3s1p1\\
resblock-4 & c260k3s1p1\\
\hline
upsample & 2x\\
\hline
conv2d-1 & c130k5s1p2\\
\hline
upsample & 2x\\
\hline
conv2d-2 & c65k5s1p2\\
conv2d-3 & c5k7s1p3\\
\hline
\end{tabular}
\end{center}
\label{tab:generator_network}
\end{table}

{\bf Discriminators:} The architecture of discriminator $D$ is shown in Table~\ref{tab:d_network}, which consists of 6 convolution layers followed by a two-head output layer: a $1\times1$ convolution layer for real-fake discrimination and a fully connected layer for modality classification.

\begin{table}[h]
\caption{Architectures of discriminator. c: number of output channels, k: kernel size, s: stride, p: padding.}
\begin{center}
\begin{tabular}{c|c}
\hline
layers & hyper-parameters\\
\hline
conv2d-1 & c64k4s2p1\\
conv2d-2 & c128k4s2p1\\
conv2d-3 & c256k4s2p1\\
conv2d-4 & c512k4s2p1\\
conv2d-5 & c1024k4s2p1\\
conv2d-6 & c2048k4s2p1\\
\hline
head-1-conv & c1k1s1p0\\
\hline
head-2-fc & c4\\
\hline
\end{tabular}
\end{center}
\label{tab:d_network}
\end{table}

The architecture of content discriminator is shown in Table~\ref{tab:d_content_network}, which has four convolution layers and a fully connected layer.

\begin{table}[h]
\caption{Architectures of content discriminator. c: number of output channels, k: kernel size, s: stride, p: padding.}
\begin{center}
\begin{tabular}{c|c}
\hline
layers & hyper-parameters\\
\hline
conv2d-1 & c256k4s2p1\\
conv2d-2 & c256k4s2p1\\
conv2d-3 & c256k4s2p1\\
conv2d-4 & c256k4s2p1\\
\hline
fc-1 & c4\\
\hline
\end{tabular}
\end{center}
\label{tab:d_content_network}
\end{table}

\section{Learning modal-independent similarity metric via image-to-image translation}

Multi-modal images from the same patient usually share similar hidden features such as the same anatomical structures (content features) but differentiate from each other in appearances such as T1 or T2 contrast in MR (style features). Thus, we hypothesize that each image can be decomposed into content features and style features, and motion information can be captured exploiting the difference in the content feature space. We extend the diverse image-to-image translation framework (DRIT++)~\cite{lee2020drit++} to learn disentangled hidden features. We train content encoder $E^c$, style encoder $E^s$ and generator $G$ to perform image translation among multiple modalities. $E^c$ maps images onto the shared content feature space and $E^s$ maps images onto the modality-specific style feature space. On the other hand, $G$ synthesizes images conditioned on both content and style features.

\subsubsection{Image translation and cycle consistency}

Suppose two different groups $\mathcal{X}_a$ and $\mathcal{X}_b$ sampled from $M$ modality groups ($\{\mathcal{X}_i\}_{i=1}^M$). Given $x_a$ and $x_b$ two image samples in these two groups, we perform cross-modality translation as follows: 

1) Extract the content and style features, i.e., $s_a= E^s(x_a, a)$, $c_a=E^c(x_a)$, $s_b= E^s(x_b, b)$, $c_b=E^c(x_b)$. 

2) Image translation between different modalities by performing image generation with swapped style features, i.e., $\hat{x}_{ab} = G(c_a, s_b, b)$, $\hat{x}_{ba}=G(c_b, s_a, a)$. 

3) Encode the translated images to reconstruct content and style features of the original images, $\hat{s}_a = E^s(\hat{x}_{ba}, a)$, $\hat{c}_a = E^c(\hat{x}_{ab})$, $\hat{s}_b = E^s(\hat{x}_{ab}, b)$, $\hat{c}_b = E^c(\hat{x}_{ba})$. 

4) Perform the second image translation by swapping the style features again to reconstruct the original images, $\hat{x}_{aba} = G(\hat{c}_a, \hat{s}_a, a)$, and $\hat{x}_{bab} = G(\hat{c}_b, \hat{s}_b, b)$. 

Eventually, we expect that $\hat{x}_{aba}$ and $\hat{x}_{bab}$ are consistent to the original images $x_a$ and $x_b$ respectively. We use L1 loss $\mathcal{L}_{\text{cyc}}$ to enforce this cycle consistency. In addition, we introduce the feature reconstruction loss for content/style features, $\mathcal{L}_{\text{rec}}^{c}$/$\mathcal{L}_{\text{rec}}^{s}$, which are the L1 loss between $c_a, c_b$/$s_a, s_b$ and $\hat{c}_a, \hat{c}_b$/$\hat{s}_a, \hat{s}_b$. Similarly, we define mono-modal reconstruction loss $\mathcal{L}_{\text{rec}}^{x}$ as the L1 loss between the reconstructed images $\hat{x}_a = G(c_a, s_a, a)$, $\hat{x}_b=G(c_b, s_b, b)$ and the original images $x_a, x_b$.

\subsubsection{Adversarial training}
To enforce encoders to learn disentangled representations, image discriminator $D$ and content discriminator $D^c$ are used via adversarial training. Given an image $x$, the output of $D$ consists of two parts: $D_{\text{src}}(x)$ and $D_{\text{cls}}(x)$, where $D_\text{src}(x)$ distinguishes synthetic images from real images and $D_\text{cls}(i|x)$ is the probability that $x$ belongs to the $i$-th modality. We use the adversarial loss in Wasserstein GAN~\cite{arjovsky2017wasserstein} for this,
\begin{equation}
\mathcal{L}_{\text{adv}}= \mathbb{E}_{x_a, x_b, a, b}[D_{\text{src}}(\hat{x}_{ab})- D_\text{src}(x_{b})].
\end{equation}

To learn style features, $D_\text{cls}$ tries to classify input images into different modalities, by minimizing cross entropy loss:
\begin{equation}
\mathcal{L}_\text{adv}^\text{cls}=-\mathbb{E}_{x_a,x_b, a,b}[\log D_\text{cls}(a|x_a)+\log D_\text{cls}(b|\hat{x}_{ab})].
\end{equation}
To learn content features and take modality independence into account, we utilize the content discriminator $D^c$ with a content adversarial loss:
\begin{equation}
\mathcal{L}_\text{adv}^c=-\mathbb{E}_{x_a, a}[\log D^c(a|c_a)].
\end{equation}

In summary, the total loss functions are as follows:
\begin{equation}
\begin{aligned}
\mathcal{L}_{G,E^s,E^c}= -\mathcal{L}_{\text{adv}}+\lambda_\text{rec}^x \mathcal{L}_{\text{rec}}^{x}
    + \lambda_\text{rec}^c \mathcal{L}_{\text{rec}}^{c}
    + \lambda_\text{rec}^s \mathcal{L}_{\text{rec}}^{s} \\
    + \lambda_\text{cyc} \mathcal{L}_{\text{cyc}}
    + \lambda_\text{adv}^\text{cls} \mathcal{L}_\text{adv}^\text{cls}
    - \lambda_\text{adv}^c \mathcal{L}_\text{adv}^c,
\end{aligned}
\end{equation}
\begin{equation}
    \mathcal{L}_{D,D^c}=\mathcal{L}_{\text{adv}}+ \lambda_\text{adv}^\text{cls} \mathcal{L}_\text{adv}^\text{cls} + \lambda_\text{gp}\mathcal{L}_\text{gp} + \mathcal{L}_\text{adv}^c,
\end{equation}
where $\mathcal{L}_{G,E^s,E^c}$ is the loss function for the generator and encoders, $\mathcal{L}_{D,D^c}$ is the loss function for discriminators, and $\mathcal{L}_{\text{gp}}$ is gradient penalty as in Wasserstein GAN~\cite{gulrajani2017improved}. $\lambda$s are the weighting coefficients.

\subsubsection{Modality-independent similarity metric}
To exploit 3D information while tackling memory usage and convergence challenges in 3D CNN, we used a multi-slice GAN strategy to train image-to-image translation, where adjacent 2D slices are concatenated along the channel dimension before fed to 2D CNN in the GAN. 

In addition, we use three orthogonal image axes: sagittal, coronal and axial slices from the volumetric data to train the network.  In the context of multi-modal image registration, we utilize the learned content encoder to extract modality independent features to measure the image similarity. The similarity metric is defined as the L2 norm in the content feature space, i.e.,
\begin{equation}
    \mathcal{L}_{\text{sim}} =\mathbb{E}_{x_a,x_b,a,b} ||E^c(x_{a}\circ\phi_{\theta_T}) - E^c(x_{b}) ||_2^2.
\end{equation}

To extend this loss to a 3D volume $x$, we randomly select $N$ slices of different axes from $x$. Let $\mathcal{I}(x)$ denote the selected set of slices, we concatenate the content features of these slices to represent the content features of the whole volume, i.e., $E^c(x)\triangleq\{E^c(y)|y\in\mathcal{I}(x)\}$. Note that the inputs and outputs of the registration ODENet are still defined in 3D.

\section{Contributors of our improvement}
We trained VM with the learned loss under the deformable setting, whose results together with other methods are listed in Table \ref{tab:loss}.

\begin{table}[h]
\caption{}
\begin{center}
\begin{tabular}{c|c|c|c}
\hline
methods & Dice & RMSE($x$)$\times10^{-2}$ & RMSE($\phi$) \\
\hline
VM & 79.4(8.7) & 8.81(3.03) & 1.61(0.69) \\
VM+I2I        & 80.1(7.8) & 8.52(2.30) & 1.26(0.31) \\
MS-ODENet(D)  & 81.6(8.1) & 6.63(2.22) & 1.11(0.18) \\
VM+learned loss & 80.5(7.7) & 7.75(2.16) & 1.18(0.20) \\
\hline
\end{tabular}
\end{center}
\label{tab:loss}
\end{table}

The learned loss (VM+learned loss) outperforms MI (VM) and I2I (VM+I2I). With the same loss, MS-ODENet achieves higher accuracy than VM+learned loss. Therefore, both the MS-ODE network and the learned metric contribute to the higher accuracy of our method over the others.

%
%
%
\bibliographystyle{splncs04}
\bibliography{egbib.bib}